\documentclass[twoside]{article}
\usepackage{aistats2019}

\usepackage{amsmath}
\usepackage{amssymb}
\usepackage{graphicx}
\usepackage{subcaption}
\usepackage{xcolor} 
\usepackage{url}
\usepackage{stfloats} 
\usepackage[round]{natbib}
\usepackage{hyperref}
\usepackage[capitalise,noabbrev]{cleveref}

\graphicspath{{figures/}}

\begin{document}

\twocolumn[
    \spectitle{ Mixed Likelihood Gaussian Process Latent Variable Model }
    \specauthor{ Samuel Murray \And Hedvig Kjellstr{\"o}m }
    \specaddress{ KTH Royal Institute of Technology, Stockholm, Sweden } 
]

\begin{abstract}
    We present the \emph{Mixed Likelihood} Gaussian process latent variable model (GP-LVM), capable of modeling data with attributes of different types. The standard formulation of GP-LVM assumes that each observation is drawn from a Gaussian distribution, which makes the model unsuited for data with e.g.\ categorical or nominal attributes. Our model, for which we use a sampling based variational inference, instead assumes a separate likelihood for each observed dimension. This formulation results in more meaningful latent representations, and give better predictive performance for real world data with dimensions of different types.
\end{abstract}

\section{INTRODUCTION}
\label{sec:introduction}
The Gaussian process latent variable model (GP-LVM)~\citep{lawrence2004gaussian} and its multiple extensions~\citep[e.g.][]{titsias2010bayesian,damianou2013deep} can be used to find low dimensional latent representations of high dimensional data. However, the original formulation assumes that the observed data are real valued with additive Gaussian noise -- or in GP terminology, that the likelihood is Gaussian. However, this assumption is often violated; medical data for example might include blood type (categorical), test results (binary), number of incidents (count), and self-reported pain level (ordinal). In this work, we will show how the GP-LVM can be extended to model data where each dimension might have a different likelihood. By allowing non-Gaussian likelihoods, we introduce a way to tailor the model to each individual problem.

As an unsupervised latent variable model, GP-LVM is capable of finding a low-dimensional representation of unlabelled data. This can be useful on its own, e.g.\ for visualization if the latent representation is two dimensional~\citep{lawrence2004gaussian}. Alternatively, it can be used as pre-processing for a supervised task.

When used for dimensionality reduction, the desired quantity in latent variable models is the posterior distribution over the latent variables. Since the posterior in GP-LVM is intractable, the first formulation used maximum likelihood estimates of the latent variables~\citep{lawrence2004gaussian}. More recently, extensions such as Bayesian GP-LVM~\citep{titsias2010bayesian} and deep GP~\citep{damianou2013deep} have been introduced that use variational inference~\citep{jordan1999introduction} to compute an approximate posterior. The key idea is to form a lower bound on the marginal likelihood using Jensen's inequality. Unfortunately, a tractable lower bound is only achieved for certain kernels and for Gaussian likelihoods. 

\citet{gal2015latent} showed that the restrictions on the choice of kernel and likelihood can be bypassed by approximating an expectation term in the lower bound using Monte Carlo techniques, and that this allows the GP-LVM to be used for categorical data. Using the same idea, we show that GP-LVM can be extended to handle any likelihood, and importantly different likelihoods for different dimensions.

Specifically, in this paper, we:
\begin{itemize}
    \item propose an extension of GP-LVM -- referred to as \emph{Mixed Likelihood GP-LVM} -- which uses a general inference to work with non-Gaussian likelihoods;
    \item present experimental results supporting our claim that previous models are special cases of Mixed Likelihood GP-LVM;
    \item demonstrate that Mixed Likelihood GP-LVM can find meaningful representations of real world datasets with attributes of different types, whereas standard GP-LVMs fail to do so, indicating that the choice of likelihood is an important part of model design.
\end{itemize}

\section{GAUSSIAN PROCESSES}
\label{sec:gp}
In regression, we have a dataset $D = \{x_n, y_n\}_{n=1}^N$ of input-output pairs, and try to find the function $f$ that maps from $x$ to $y$. We commonly assume additive Gaussian observation noise, i.e. that\
\begin{equation}
    y_n = f(x_n) + \epsilon_n, \; \epsilon_n \sim ~ \mathcal{N}(0, \, \sigma^2).
\end{equation}
Rather than specifying a parametric form, we can place a (zero-mean) Gaussian process prior over $f$ and compute its posterior given the observed data. We denote this as
\begin{equation}
    f(\cdot) \sim \mathcal{GP}(0, \, k(x, \, x')),
\end{equation}
where the prior only depends on the covariance function $k(x, x')$ -- also known as kernel -- which computes the covariance of function values between two input locations. Hence, the kernel encapsulates all modeling decisions, and it is well-known that an appropriate choice of kernel can greatly improve the performance of GPs~\citep{rasmussen2006gaussian}. One of the nice properties of GPs is that the predictive posterior at a new location $x_*$ can be computed in closed form by integrating out the function $f$.

Gaussian processes can also be used for other tasks than regression, for example binary classification. However, whereas in standard regression $f$ was the noise-free output, here $f$ should be seen as a latent function that maps $x$ to a parameter of the likelihood. In the case of logistic regression, the likelihood can be written as $y|x \sim Ber(\sigma(f(x)))$, where $\sigma(\cdot)$ is a sigmoid function ensuring that the parameter of the Bernoulli distribution is between $0$ and $1$. Similarly, in Poisson regression we can write the likelihood as $y|x \sim Pois(\exp(f(x)))$. In fact, for any distribution $D$ parameterized by $\theta$, the likelihood can be written as $y|x \sim D(h(f(x)))$, where $h$ is function that maps $f$ to allowed values of $\theta$. We refer to $h$ as an inverse link function, borrowing from the literature of generalized linear models~\citep{mccullagh1989generalized}. The problem with using a non-Gaussian likelihood is that most quantities of interest (e.g.\ the predictive posterior) no longer have closed form expressions. Instead one must resort to approximation methods.

\section{MIXED LIKELIHOOD GP-LVM}
\label{sec:mlgplvm}
How can Gaussian processes be used in unsupervised learning? The key idea in GP-LVM~\citep{lawrence2004gaussian, titsias2010bayesian} is to keep the same model, but view the input $X$ as latent variables. That is, we posit a generative model of our observed data, which are viewed as the output of a Gaussian process. By placing a prior on the unobserved inputs, we can compute the posterior of the inputs (latent variables) given the outputs (observed data). By forcing the input to be of lower dimension than the output, we can find a low-dimensional representation of the observed data. Importantly, in this work we do not constrain ourselves to Gaussian likelihoods, which allows to us to work with data of any type.

Formally, let $\mathcal{S}_1, \dots, \mathcal{S}_D$ be $D$ sets and consider the Cartesian product $\mathcal{S} = \prod_{d=1}^D \mathcal{S}_d$. Each set $\mathcal{S}_d$ corresponds to one observed dimension, and can for example be $\mathbb{R}$ (real), $\mathbb{N}$ (non-negative integer), $\{0, 1\}$ (binary) or $\{C_0, \dots, C_K\}$ (categorical). Let $\mathbf{Y} = (\mathbf{y}_1, \dots , \mathbf{y}_N)^T \in \mathcal{S}^N$ be the observed data, i.e.\ each data point $\mathbf{y}_n$ lies in $\mathcal{S}$ and $N$ is the total number of data points. The goal is to find a latent representation $\mathbf{X} \in \mathbb{R}^{N \times Q}$ of the data, where $Q < D$. We place a Gaussian prior over $\mathbf{X}$, and let $f(\mathbf{X)} = \mathbf{F} \in \mathbb{R}^{N \times D}$ be latent function values with $f$ being drawn from a Gaussian process. 

In the standard formulation of GP-LVM, only Gaussian likelihoods are used, i.e.\ $y_{nd}|f_{nd} \sim \mathcal{N}(f_{nd}, \sigma^2)$. In a similar vein, to model categorical data, \citet{gal2015latent} use the softmax function to map from $f_{nd}$ to the probabilities of a categorical distribution. Conversely, we assume that each dimension can have a different likelihood, and thus have that $y_{nd}$ is drawn from any distribution with parameters determined by $f_{nd}$ and an inverse link function $h_d(\cdot)$. In general, $f_{nd}$ can be a vector if the distribution has multiple parameters -- as is the case with the categorical distribution -- but for ease of notation we make no such distinction.

As in Bayesian GP-LVM, introduced by~\citet{titsias2010bayesian}, we augment our model with $M$ auxiliary inducing points to facilitate inference. Specifically, we assume that $\mathbf{U} \in \mathbb{R}^{M \times D}$ are function values of $f$ evaluated at the inducing inputs $\mathbf{Z} \in \mathbb{R}^{M \times Q}$. Note that $\mathbf{U}$ and $\mathbf{Z}$ lie in the same spaces as $\mathbf{F}$ and $\mathbf{X}$ respectively, and that the same mapping $f$ is used. The resulting augmented model is
\begin{equation}
    p(\mathbf{Y}, \mathbf{F}, \mathbf{U}, \mathbf{X}) = p(\mathbf{Y}|\mathbf{F}) p(\mathbf{F}|\mathbf{U}, \mathbf{X}) p(\mathbf{U}) p(\mathbf{X}).
\end{equation}
Since we condition our latent function values on the inducing outputs, these should ideally summarize the mapping $f$ at the positions of our latent points, which is achieved by optimizing the placement of $Z$. Note however that by integrating out the inducing points we recover the original model, and thus they do not affect the marginal likelihood $p(\mathbf{Y})$. 

We use the RBF kernel with automatic relevance determination (ARD). It is a popular choice in GP-LVM due to its ability to automatically determine the dimensionality of the latent space~\citep{titsias2010bayesian,damianou2013deep}. Given a variance $\sigma_{rbf}^2$ and $Q$ number of inverse lengthscales $\gamma_q$, the covariance between $\mathbf{x}$ and $\mathbf{x}'$ is computed as $k(\mathbf{x}, \mathbf{x}') = \sigma_{rbf}^2 \exp [\sum_q \gamma_q (x_q - x_q')^2]$. The inverse lengthscales determine the relevance of each dimension, with a high value being interpreted as high relevance, thus motivating the name ARD RBF.

Let $K_{xx}$ be the covariance matrix such that $K_{xx}[i, j] = k(\mathbf{x}_i, \mathbf{x}_j)$, and similarly for $K_{zz}$, $K_{xz}$ and $K_{zx}$. Our generative process can be written as
\begin{align}
    x_{nq} &\sim \mathcal{N}(0, I), \\
    \begin{bmatrix}
        \mathbf{f}_d \\
        \mathbf{u}_d
    \end{bmatrix}
        &\sim \mathcal{N}
            \Big(0, 
            \begin{bmatrix} 
                K_{xx} & K_{xz} \\
                K_{zx} & K_{zz}
            \end{bmatrix}
            \Big), \\
    y_{nd}|f_{nd} &\sim \text{Likelihood}_d(h_d(f_{nd})),
\end{align}
where $h_d(\cdot)$ is the deterministic inverse link function mapping $f_{nd}$ to allowed parameter values of $\text{Likelihood}_d$. 

All exponential family distributions have a canonical (inverse) link function: for a Gaussian likelihood, $h(f_{nd}) = [f_{nd}, \sigma_d^2]$ where the variance $\sigma_d^2$ may be optimized or kept fixed; for binary data and a Bernoulli likelihood, $h(f_{nd}) = \sigma(f_{nd})$ where $\sigma(\cdot)$ is the logistic function; for categorical data, $h(f_{nd}) = \text{softmax}(f_{nd})$; for count data and a Poisson likelihood, $h(f_{nd}) = \exp(f_{nd})$. In principle, any likelihood and inverse link function can be used as long as we can evaluate the log likelihood and compute the gradient of the log likelihood with respect to $f_{nd}$. We also note that by using the canonical link functions, our model is the same as Bayesian GP-LVM~\citep{titsias2010bayesian} if each likelihood is Gaussian, and the same as Categorical Latent Gaussian Process (CLGP)~\citep{gal2015latent} if each likelihood is a categorical distribution.

\section{INFERENCE}
\label{sec:inference}
In Bayesian modelling, the quantity of interest is the marginal likelihood $p(\mathbf{Y})$, since it is a measure of how well the model describes the data. With this quantity, we could also compute the posterior over the latent inputs as $p(\mathbf{X}|\mathbf{Y}) = p(\mathbf{Y}|\mathbf{X}) p(\mathbf{X}) / p(\mathbf{Y})$. Unfortunately, the marginal likelihood (and thus also the posterior) is intractable. The standard procedure in recent GP-LVM work, first proposed by \citet{titsias2010bayesian}, is to use variational inference and maximize a lower bound of the marginal likelihood. The first step is to introduce the following approximate posterior over the latent variables:
\begin{equation}
    q(\mathbf{F}, \mathbf{U}, \mathbf{X}) = q(\mathbf{X}) q(\mathbf{U}) p(\mathbf{F} | \mathbf{U}, \mathbf{X}).
\end{equation}
The variational distributions $q(\mathbf{X})$ and $q(\mathbf{U})$ are taken to be Gaussian distributions, factorized over the dimensions:
\begin{align}
    q(\mathbf{X}) &= \prod_{q=1}^Q \mathcal{N}(\boldsymbol{\mu}_q^X, \Sigma_q^X), \\
    q(\mathbf{U}) &= \prod_{d=1}^D \mathcal{N}(\boldsymbol{\mu}_d^U, \Sigma_d^U).
\end{align}

\subsection{Evidence lower bound}
\label{sec:inference:elbo}

By using the results of \citet{titsias2010bayesian} and \citet{gal2015latent}, we can get the lower bound $\mathcal{L} \leq \log p(\mathbf{Y})$ by applying Jensen's inequality. This quantity is referred to as the evidence lower bound (ELBO) and has the form
\begin{align}
\label{eq:elbo}
    \begin{split}
        \mathcal{L} = 
            &- \sum_{d=1}^D \text{KL}(q(\mathbf{u}_d)\,\|\, p(\mathbf{u}_d))
            - \sum_{q=1}^Q \text{KL}(q(\mathbf{x}_q)\,\|\, p(\mathbf{x}_q)) \\
            &+ \sum_{n=1}^N \sum_{d=1}^D \langle \log p(y_{nd} | f_{nd})\rangle_{q(\mathbf{x}_n) q(\mathbf{u}_d) p(f_{nd}|\mathbf{u}_d, \mathbf{x}_n)},
    \end{split}
\end{align}
where $\langle \cdot \rangle$ denotes an expectation. As shown by \citet{gal2015latent}, the last term can be approximated using Monte Carlo sampling, since
\begin{align}
\label{eq:mc_expectation}
    \begin{split}
        \langle \log p(y_{nd} | f_{nd})&\rangle_{q(\mathbf{x}_n) q(\mathbf{u}_d) p(f_{nd}|\mathbf{u}_d, \mathbf{x}_n)} \\
        &\approx \frac{1}{T} \sum_{t=1}^T \log p(y_{nd} | f_{nd}^t),
    \end{split}
\end{align}
where $\mathbf{f}_d^t = (f_{1d}^t, \cdots, f_{Nd}^t)$ is generated by sampling $\mathbf{u}_d^t$ and $\mathbf{X}^t$ from their variational distributions, and subsequently $\mathbf{f}_d^t$ from $p(\mathbf{f}_d | \mathbf{u}_d^t, \mathbf{X}^t)$. If we assume that $q(\mathbf{X})$ factorizes over data points -- i.e.\ if $\Sigma_q^X$ is taken to be diagonal -- the KL term can be computed independently across data points.

By reparameterizing $q(\mathbf{X})$, $q(\mathbf{U})$ and $p(\mathbf{F} | \mathbf{U}, \mathbf{X})$, we can compute the gradients with respect to the parameters in the sampling procedure. The reader is referred to \citet{gal2015latent} for the exact expressions.

\subsection{Partially observed data}
\label{sec:inference:partially_observed_data}

Our model naturally works with partially observed data. Assume that some data points have missing attributes, and that we still want to use all data to form the latent space, possibly in order to impute the missing values. First, we note that the observed values only affect the ELBO in the third term of \cref{eq:elbo}, and that through our sampling this factorizes over dimensions and data points. We can thus treat all data points the same way, without computing the log probability of any missing value. Note however that the latent inputs will be optimized in the same way, and thus we can find latent inputs of noisy data points. Subsequently we can compute the posterior of the missing values given the latent inputs, conditioned on the inducing points. However, since this is not tractable in general when using non-Gaussian likelihoods, we resort to sampling, either using \cref{eq:mc_expectation}, or by keeping $\mathbf{X}$ and $\mathbf{U}$ fixed to their mean values.

\subsection{Optimization}
\label{sec:inference:optimization}

We optimize all the parameters using gradient decent to maximize the ELBO. Since we only sample from Gaussian distributions, which are fully reparametrizable, we can compute the gradients through our sampling procedure. We implement our model using Tensorflow Probability~\citep{dillon2017tensorflow}, which allows us to simply express the ELBO and use automatic differentiation to compute all the gradients. Tensorflow Probability has implementations for Monte Carlo sampling and a multitude of probability distributions which we can use as likelihood functions. Notably, the only requirements for a distribution to be used as a likelihood are that the log probability can be evaluated, and that the gradients of the log probability with respect to its parameters can be computed.

The parameters to optimize over are the kernel parameters ($\sigma_{rbf}^2$, $\gamma_q$), the parameters of $q(\mathbf{X})$ and $q(\mathbf{U})$ ($\boldsymbol{\mu}_q^X$, $\Sigma_q^X$, $\boldsymbol{\mu}_d^U$, $\Sigma_d^U$), the positions of the inducing points $\mathbf{Z}$, and any free parameters of the likelihood functions ($\sigma_d^2$ for Gaussian likelihoods). Since $\sigma_d^2$, $\sigma_{rbf}^2$ and $\gamma_q$ must be positive, we optimize the log of these quantities; since $\Sigma_q^X$ and $\Sigma_d^U$ must be symmetric and positive semi definite, we optimize the lower triangular matrices of the Cholesky decompositions instead. We use RMSProp to jointly optimize all parameters, with the loss being the negative ELBO. Unless otherwise stated, the same initialization is used in all experiments. The mean of the latent inputs $\boldsymbol{\mu}_q^X$, the inducing inputs $\mathbf{Z}$, and the mean of the inducing outputs $\boldsymbol{\mu}_d^U$ are all initialized randomly from a zero mean, unit variance normal distribution. We initialize the covariance matrices ($\Sigma_q^X$ and $\Sigma_d^U$) as identity matrices. The initial value of the kernel variance $\sigma_{rbf}^2$ is $1$, and all inverse lengthscales $\gamma_d$ are set to $0.5$. In the case of Gaussian likelihoods, the variance $\sigma_d^2$ of each dimension has an initial value of $0.1$. We use 50 inducing points, 10 samples to approximate the expectation, an initial learning rate of $10^{-3}$, and run until the ELBO has converged.

\section{EXPERIMENTS}
\label{sec:experiments}
Since our model is a generalization of Bayesian GP-LVM~\citep{titsias2010bayesian} and Categorical Latent Gaussian Process (CLGP)~\citep{gal2015latent} (as described in \cref{sec:mlgplvm}) we first conduct two experiments to establish that it works as intended in the special cases where all likelihoods are Gaussian or categorical. We subsequently demonstrate that it works well for data with different types -- settings where previous models fail. Since those models are not designed to work with other likelihoods, the goal is not to criticize them, but rather to show that our model can be used to tackle new types of problems. We use labeled datasets when evaluating our model as it enables us to reason about the quality of the latent space. Note however that the labels are never observed by the model, and that in real world setting we would in general not have access to any labels. The results are summarized in \cref{tab:results}.

\begin{table*}[t!]
    \centering
    \caption{Results from different experiments. The CLGP is only evaluated on Binary Alpadigits, since it requires categorical data. 1NN errors measures missclassifications using one nearest neighbor; 1NN RMSE measures the root mean square error; the log likelihood is computed with base $e$ whereas the log perplexity is computed with base $2$; the last row shows the log likelihood evaluated only on data with missing gender. An up (down) arrow indicates that higher (lower) values are better; the best value for each experiment is in \textbf{bold}. $^{(*)}$ taken from \citet{titsias2010bayesian}. $^{(**)}$ taken from \citet{gal2015latent}.}
    \label{tab:results}
    \begin{tabular}{l|l|c|c|c|c}
        \textbf{Dataset}    & \textbf{Measure}                          & \textbf{MLGPLVM}  & \textbf{BGPLVM}       & \textbf{CLGP}     & \textbf{PCA}  \\ \hline
        Oilflow             & 1NN errors $(\downarrow)$                 & $8$               & $\mathbf{3} ^{(*)}$   & --                & $26 ^{(*)}$   \\ \hline
        Binary Alphadigits  & Test log perplexity $(\downarrow)$        & $\mathbf{0.64}$   & --                    & $\sim 0.7 ^{(**)}$& --            \\ \hline
        Cleveland           & 1NN errors (binary) $(\downarrow)$        & $\mathbf{69}$     & $125$                 & --                & $122$         \\ \hline
        Cleveland           & 1NN errors (0-4) $(\downarrow)$           & $\mathbf{134}$    & $176$                 & --                & $179$         \\ \hline
        Cleveland           & Test log likelihood $(\uparrow)$          & $\mathbf{-1925}$  & $-7776$               & --                & --            \\ \hline
        Abalone             & 1NN RMSE $(\downarrow)$                   & $\mathbf{3.20}$   & $3.64$                & --                & $3.39$        \\ \hline
        Abalone             & Test log likelihood $(\uparrow)$          & $366$             & $\mathbf{4490}$       & --                & --            \\ \hline
        Abalone             & Test log likelihood (gender) $(\uparrow)$ & $\mathbf{34}$     & $-4397$               & --                & --            \\
    \end{tabular}
\end{table*}

\subsection{Oilflow}
\label{sec:experiments:oilflow}

In order to confirm that our inference is valid, we begin by comparing our model with Bayesian GP-LVM when only Gaussian likelihoods are used. The Oilflow dataset~\citep{bishop1993analysis} has long been used in GP-LVM literature to evaluate models~\citep{lawrence2004gaussian,titsias2010bayesian,gal2014variational,damianou2016variational}, and is thus a natural starting point. The dataset consists of $1000$ points with each point belonging to one of three classes. The data are 12-dimensional and real-valued. Though the meaning of the classes is not relevant in this experiment, we know from previous work that the data can be reduced to two dimensions in such a way that the classes are nicely separated.

\begin{figure}[!b]
    \begin{subfigure}{0.5\columnwidth}
        \includegraphics[width=\linewidth]{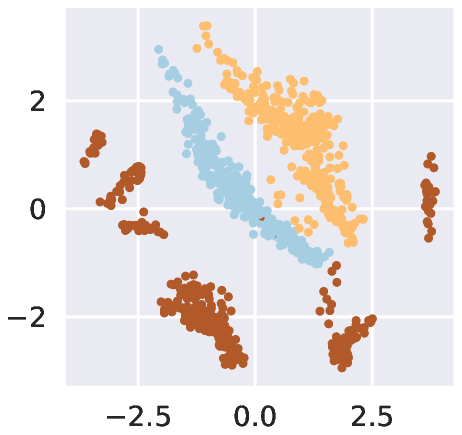}
        \caption{MLGPLVM}
        \label{fig:oilflow_mlgplvm}
    \end{subfigure}%
    \begin{subfigure}{0.5\columnwidth}
        \includegraphics[width=\linewidth]{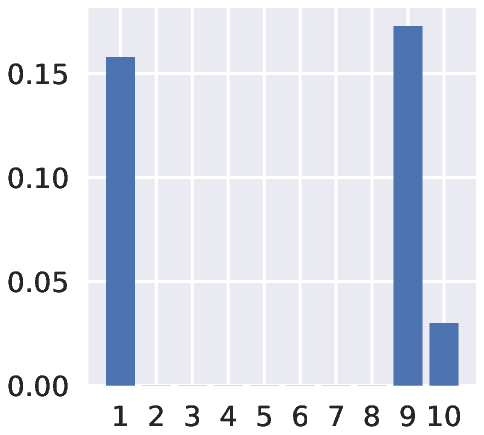}
        \caption{Inverse lengthscales}
        \label{fig:oilflow_gamma}
    \end{subfigure}
    \caption{Results on the Oilflow dataset for Mixed Likelihood GP-LVM. Each of the three classes is represented by a color. (a) shows the latent space when keeping the two most dominant dimensions ($9$ and $1$). (b) shows the inverse lengthscales from the ARD RBF kernel.}
    \label{fig:oilflow}
\end{figure}

Since all dimensions are real-valued, we use only Gaussian likelihoods. This way, our model is the same as Bayesian GP-LVM. We use a similar setup as \citet{titsias2010bayesian}, with 50 inducing points and 10 latent dimensions, but instead of initializing the latent input locations $\mathbf{X}$ with PCA, we sample them from a zero mean, unit variance normal distribution. The inverse lengthscales are displayed in \cref{fig:oilflow_gamma}, and it can be noted that all but three dimensions are switched off. To visualize the approximate posterior, we keep only the two most dominant dimensions ($9$ and $1$), and plot the mean values in \cref{fig:oilflow_mlgplvm}. Arguably, the separation of the classes is not as good as in \citet{titsias2010bayesian}, however it is still of high quality. We also look at the nearest neighbour of each point and see if they are from the same class (i.e.\ the $1$ nearest neighbour classification error). Our model has $8$ misclassified points, whereas Bayesian GP-LVM has $3$, and simply using PCA with two dimensions gives an error of $26$. 

The fact that Mixed Likelihood GP-LVM performs slightly worse than Bayesian GP-LVM is actually not surprising; even though the \emph{model} is the same, the inference is different. Bayesian GP-LVM has an analytical expression of the lower bound that explicitly assumes Gaussian likelihoods -- in contrast, our approach uses Monte Carlo estimates, which works for any likelihood but comes at the cost of noisy gradients. Thus, if all likelihoods are Gaussian, it is better to use the standard formulation rather than our more general inference. Still, the results are comparable, which shows that our inference is indeed valid.

\subsection{Binary Alphadigits}
\label{sec:experiments:binary_alphadigits}

We deviate slightly from \citet{gal2015latent} in how we initialize and optimize our model. Most notably, in CLGP different sets of parameters are updated iteratively in turns, whereas we optimize all parameters simultaneously. We go on to verify that this does not affect the performance negatively.

\begin{figure}[!b]
    \centering
    \includegraphics[width=\linewidth]{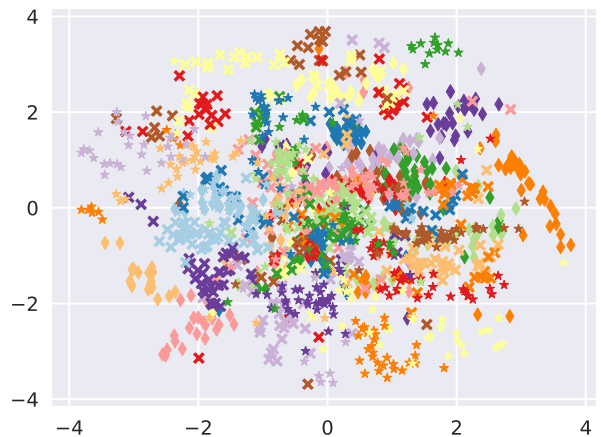}
    \caption{The two-dimensional latent representation of Mixed Likelihood GP-LVM on the Binary Alphadigits dataset. Each color-marker combination represents one of the 36 classes.}
    \label{fig:alphadigits}
\end{figure}

The Binary Alphadigits dataset\footnote{\url{https://cs.nyu.edu/~roweis/data.html}, Accessed: 11/9 2018.} consists of binary images of size $20 \times 16$. There are 36 classes, one for each digit and letter, and each class has 39 images. We reproduce the results of \citeauthor{gal2015latent}, in order to show that our model has similar performance as CLGP in the case where all likelihoods are categorical. In fact, we use Bernoulli distributions instead of categorical distributions with two classes for each likelihood, but this turns out to be identical since CLGP contrains the first input of the softmax function to be $0$. The setup of the experiment is the same, i.e.\ we downsample the images to $10 \times 8$ and flatten them to binary vectors of size $80$. From each class, we randomly select 9 images (referred to as the test set), and from each of these remove 20\% of the pixels. All images are then given as input to our model, which finds a two-dimensional latent representation without any information of the classes. The representation, shown in \cref{fig:alphadigits}, separates the classes fairly well, and bears clear similarities to that of CLGP. However, to get a quantitative measure, we compute the log perplexity on the noise-free test set. This measures how probable the test points are under our model, i.e.\ it tells us how well the model can predict the missing values. We get a log perplexity of $0.64$ -- CLGP has a reported value of $0.7$ -- which shows that the model is not dependent on specific hyperparameters, and that optimizing the parameters separately is not required.

\subsection{Cleveland}
\label{sec:experiments:cleveland}

Next, we test our model on data with dimensions of different types. Our claim is that, by specifying an appropriate likelihood for each dimension, our model is able to find a better latent representation of such data.

\begin{figure}[t]
    \begin{subfigure}{0.5\columnwidth}
        \includegraphics[width=\linewidth]{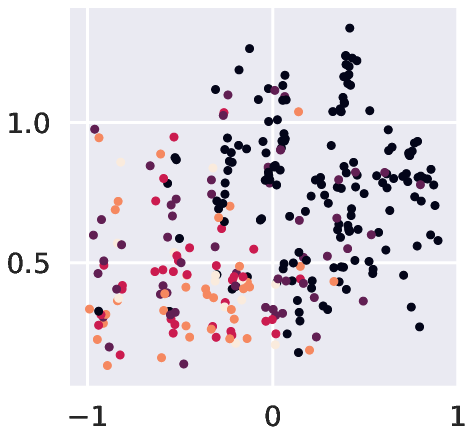}
        \caption{MLGPLVM}
        \label{fig:cleveland_mlgplvm}
    \end{subfigure}%
    \begin{subfigure}{0.5\columnwidth}
        \includegraphics[width=\linewidth]{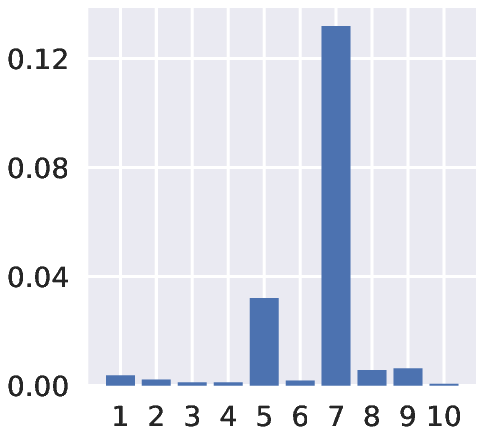}
        \caption{Inverse lengthscales}
        \label{fig:cleveland_gamma}
    \end{subfigure}
    \begin{subfigure}{0.5\columnwidth}
        \includegraphics[width=\linewidth]{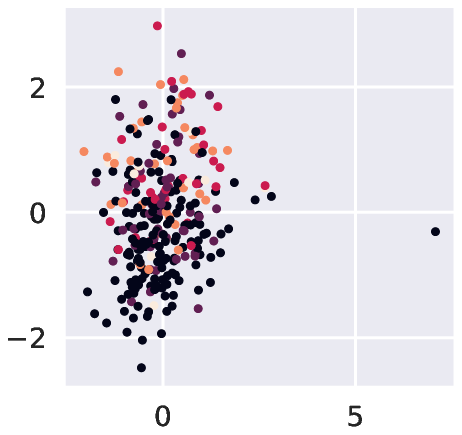}
        \caption{BGPLVM}
        \label{fig:cleveland_bgplvm}
    \end{subfigure}%
    \begin{subfigure}{0.5\columnwidth}
        \includegraphics[width=\linewidth]{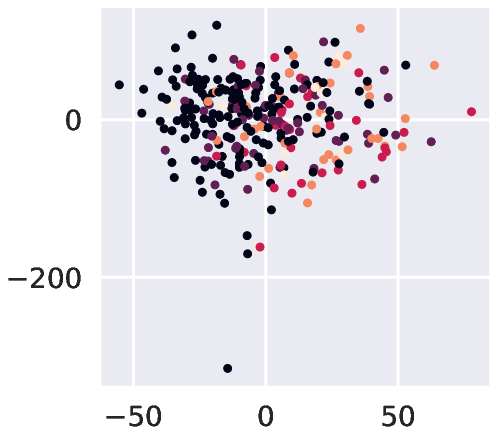}
        \caption{PCA}
        \label{fig:cleveland_pca}
    \end{subfigure}
    \caption{Results on the Cleveland dataset. (a) shows the latent space of Mixed Likelihood GP-LVM when keeping the two most dominant dimensions ($7$ and $5$). (b) shows the inverse lengthscales from the ARD RBF kernel. (c) shows the latent space of Bayesian GP-LVM. (d) shows the visualization of PCA. Black dots are used for class $0$ (no presence), and lighter colors for higher values ($1$-$4$).}
    \label{fig:cleveland}
\end{figure}

The Cleveland database~\citep{detrano1989international} is a medical dataset with 297 examples available through the UCI Machine Learning Repository~\citep{dua2017uci}. It consists of 13 attributes, including age, gender, blood pressure and cholesterol values, some of which are real valued, some binary and some categorical. The dataset is mostly used in supervised learning, where the goal is to predict the presence of heart disease, an integer ranging from $0$ (no presence) to $4$. However, it is common to treat it as a binary classification task, i.e.\ predict $0$ or $\ge 1$. We use all data, except the label, and for each dimension specify a likelihood function. Provided in the metadata is that 5 dimensions are real valued, 3 are binary and 5 are categorical, taking 3 or 4 possible values. We use one-hot encoding for the categorical variables, which when flattened makes our data points be vectors of size $25$. We thus specify our model with 5 Gaussian, 3 Bernoulli and 5 categorical likelihoods, corresponding to the different types. We use a 10-dimensional latent space, and visualize the two dimensions with largest inverse lenghtscales in \cref{fig:cleveland_mlgplvm}. Each data point is colored according to its true label, with darker colors representing smaller values (black is $0$). We note from \cref{fig:cleveland_gamma} that the model views two dimensions as much more relevant than the others, and that these dimensions seem to arrange the data points according to the label, albeit not perfectly. However, we emphasize that this is done in a completely unsupervised setting.

The same experiment is repeated with Bayesian GP-LVM and PCA. The former uses the same number of latent dimensions and an ARD RBF kernel, where we again visualize the two dimensions with largest inverse lengthscales. With PCA, we pick the two dimensions with the largest eigenvalues. Both models fail to separate the data by class labels, as can be seen in \cref{fig:cleveland_bgplvm,fig:cleveland_pca}. For each model, we compute the number of nearest neighbor errors using only two dimensions, both when using the ordinal and the binary labels. The results in \cref{tab:results} show that our model achieves a lower error in both cases, and that Bayesian GP-LVM and PCA perform similarly to each other.

Additionally, we investigate how well our model can predict missing values for this data set. For this, we randomly select 20\% of the data, and for each of these points remove 2 attributes at random. We retrain the models using all data, with some points now being partially observed. We find a two-dimensional latent space, and use the latent representations of the test points to compute the log likelihood on the test data. Mixed Likelihood GP-LVM gets a value of $-1925$, significantly better than Bayesian GP-LVM which gets $-7776$. This shows that the choice of likelihood is important; both PCA and Bayesian GP-LVM are misspecified in that they posit a Gaussian likelihood for each dimension, which is not appropriate for categorical data.

\subsection{Abalone}
\label{sec:experiments:abalone}

Finally, we evaluate the performance of Mixed Likelihood GP-LVM on a larger regression dataset. This is to see if it can find meaningful latent representations when the data points to not belong to distinct, unobserved categories, and to confirm that it works with more than a few hundred observations.

\begin{figure}[t]
    \begin{subfigure}{0.47\columnwidth}
        \includegraphics[width=\linewidth]{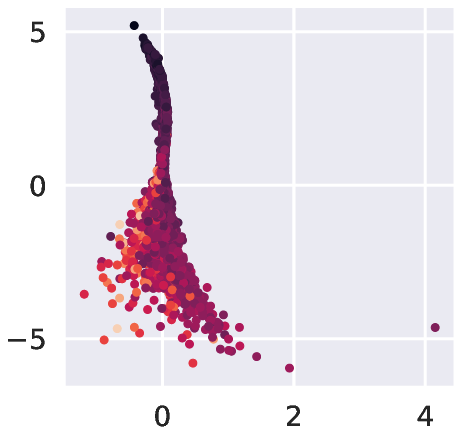}
        \caption{MLGPLVM}
        \label{fig:abalone_mlgplvm}
    \end{subfigure}%
    \begin{subfigure}{0.49\columnwidth}
        \includegraphics[width=\linewidth]{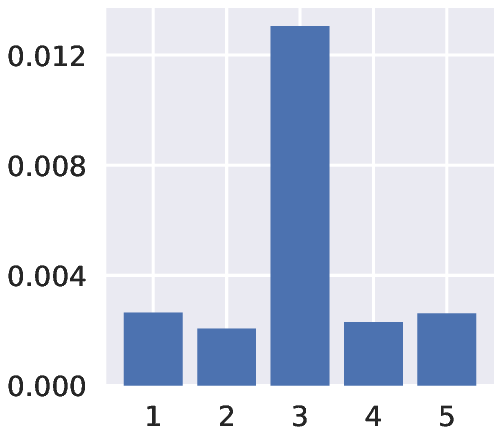}
        \caption{Inverse lengthscales}
        \label{fig:abalone_gamma}
    \end{subfigure}
    \begin{subfigure}{0.47\columnwidth}
        \includegraphics[width=\linewidth]{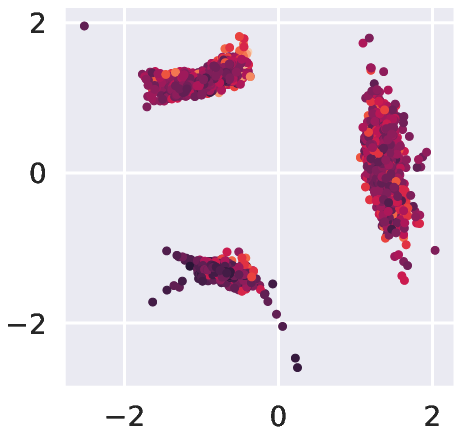}
        \caption{BGPLVM}
        \label{fig:abalone_bgplvm}
    \end{subfigure}%
    \begin{subfigure}{0.47\columnwidth}
        \includegraphics[width=\linewidth]{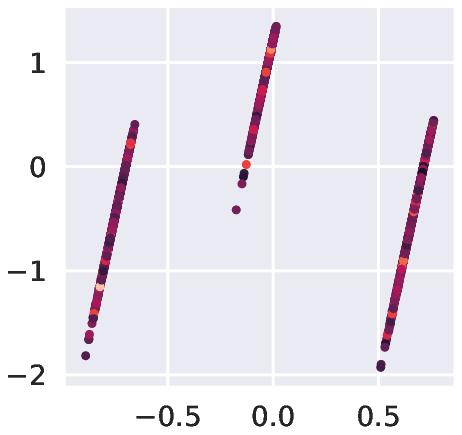}
        \caption{PCA}
        \label{fig:abalone_pca}
    \end{subfigure}
    \caption{Results on the Abalone dataset. (a) shows the latent space of Mixed Likelihood GP-LVM when keeping the two most dominant dimensions ($3$ and $1$). (b) shows the inverse lengthscales from the ARD RBF kernel. (c) shows the latent space of Bayesian GP-LVM. (d) shows the visualization of PCA. Colors correspond to age, with
    lighter colors representing higher age.}
    \label{fig:abalone}
\end{figure}

The Abalone dataset~\citep{nash1994population} consists of 4177 data points, and the goal is to predict the age of abalones from physical measurements. There are eight attributes -- seven length and weight measurements and one categorical attribute indicating male, female or infant (referred to as gender from now on). We use all attributes except the label, with a Gaussian likelihood for the physical measurements and a categorical likelihood for the gender. We find a five-dimensional latent representation of the data, and keep the two most dominant dimensions, according to the inverse lengthscales in \cref{fig:abalone_gamma}. The result is shown in \cref{fig:abalone_mlgplvm}, with each point being colored by its age, lower age having darker color. The most dominant dimension, represented by the vertical axis, seems to captures a trend of increasing age. It is interesting that the second dimension shows larger variability in the bottom part of the figure. We hypothesize that this comes from the fact that younger specimens naturally have less varying sizes, in absolute numbers. 

We run the same experiment with Bayesian GP-LVM, using five latent dimensions and an ARD RBF kernel, and with PCA, extracting the two dimensions with highest eigenvalues. As can be seen in \cref{fig:abalone_pca}, PCA produces three parallel lines on which all data lie. By inspection, we find that these correspond to the gender value, i.e.\ all female data points lie on one line and similarly for male and infant. It is easy to see why this happens: the categorical variables are either $0$ or $1$, whereas most of the other values are distributed between $0$ and $1$. Therefore the categorical attribute seems to have high variability, and thus high information content due to the Gaussian assumption of PCA. Bayesian GP-LVM yields a similar plot, where again the data have been split into three separate groups. However, since the mapping is non-linear, it can spread the points rather than placing them on straight lines. It should be noted that similar results are obtained if the gender is encoded densely $(0, 1, 2)$ instead of with one-hot vectors, with the difference that the clusters would be ordered according to the categorical value. We consider one nearest neighbour regression for all points, and compute the root mean squared error given the true labels. Mixed Likelihood GP-LVM achieves the lowest value of $3.20$, compared to $3.77$ and $3.39$ for Bayesian GP-LVM and PCA, respectively.

As with the Cleveland dataset, we compare how well the models can predict missing values. We remove one attribute at random from 20\% of the data, and fit the models using two latent dimensions. We then compute the log likelihood of the data points with missing values. As can be seen in \cref{tab:results}, Bayesian GP-LVM achieves a better score than our model. However, if we only consider the data points where the gender attribute is missing, we get the reverse result. This indicates that although there might be some merit in separating the data into clusters by gender, this prevents the model from accurately predicting missing gender information. In other words, the latent representation of Bayesian GP-LVM has the property that the nearest neighbour of all points are of the same gender, but this is achieved \emph{using} the gender variable, rather than correlating it with other attributes. Thus when it sees a new data point with missing gender, it can not assign this to either of the classes and in that sense, we argue, it overestimates the importance of the categorical value.

\section{RELATED WORK}
\label{sec:related_work}
There has been much work on extending GP-LVM to new types of data since its formulation by \citet{lawrence2004gaussian}. The two most relevant advances for this work, Bayesian GP-LVM~\citep{titsias2010bayesian} and CLGP~\citep{gal2015latent} have already been discussed. The model has also been extended to work for temporal data~\citep{wang2006gaussian,damianou2011variational}, where the latent variables are seen as draws from a GP over time. It has also been shown how the model can be modified to enable training to be done in minibatches~\citep{hensman2013gaussian}. Another line of work has been on deep GPs~\citep{damianou2013deep}. The idea is to feed the output of a GP as the input to another GP, where the first input layer is either observed (supervised) or latent (unsupervised). 

Gaussian processes have also been extensively used in supervised learning, and different methods have been proposed to handle non-Gaussian likelihoods~\citep{rasmussen2006gaussian,chu2005gaussian}. There has been an interesting work by \citet{bonilla2016generic}, presenting a general inference scheme similar to ours for any likelihood. However, they do not consider unsupervised tasks, or multiple outputs with separate likelihoods.

Also related to our approach, there has been much work outside the field of Gaussian processes. General variational inference algorithms have been proposed which require little adjustment for new models~\citep{ranganath2014black}. This is used in Deep Exponential Families (DEFs)~\citep{ranganath2015deep}, a class of deep latent variable models where the output of each layer is modeled with an exponential family distribution. Knowledge about the data is explicitly used in model design through the choice of likelihood, and inverse link functions are used in the same way as descriped here. Naturally, this has been a source of inspiration for our work. However, DEFs differ from our model in that a parametric form is used for the latent functions, and that different likelihoods are not used for the observed data.

\section{CONCLUSIONS}
\label{sec:conclusions}
We have introduced a general model capable of finding a shared continuous latent space of data with dimensions of different types. Our model is an extension of the Gaussian process latent variable model, in that it allows for different dimensions to have different likelihoods. Many real world datasets, including medical records, have attributes with distinct types, and by specifying appropriate likelihoods, we can account for this in a principled way. We show on diverse datasets that this leads to more informative latent spaces than obtained with a standard GP-LVM. Further, we show that our model can be used with partially observed data, and that using non-Gaussian likelihoods results in better predictive performance when attributes are not real-valued.

It would be interesting to extend this work with ideas from other Gaussian process models. The framework naturally allows to be used in supervised settings, by treating the input as fully observed rather than optimizing the posterior over it. Alternatively, the input could be noisy, with Gaussian prior distributions. In a similar vein, this model could be extended to work with temporal data, by including a dynamical prior over the input.

\subsubsection*{Acknowledgements}
This work was partially supported by the Wallenberg AI, Autonomous Systems and Software Program (WASP) funded by the Knut and Alice Wallenberg Foundation.

\bibliographystyle{abbrvnat}
\bibliography{refs}

\end{document}